\documentclass{ieeeaccess}
\usepackage{cite}
\usepackage{amsmath,amssymb,amsfonts}
\usepackage{algorithmic}
\usepackage{graphicx}
\usepackage{caption}%
\usepackage{url}
\usepackage{textcomp}
\def\BibTeX{{\rm B\kern-.05em{\sc i\kern-.025em b}\kern-.08em
    T\kern-.1667em\lower.7ex\hbox{E}\kern-.125emX}}
\begin{document}
\history{Date of publication xxxx 00, 0000, date of current version xxxx 00, 0000.}
\doi{10.1109/ACCESS.2017.DOI}

\title{Power Plays: Unleashing Machine Learning Magic in Smart Grids}

\author{\uppercase{Abdur Rashid}\authorrefmark{1}, 
\uppercase{Parag Biswas}\authorrefmark{1}, \uppercase{abdullah al masum}\authorrefmark{1}, 
\uppercase{MD Abdullah Al Nasim}\authorrefmark{2}, 
\uppercase{Kishor Datta Gupta}\authorrefmark{3}}

\address[1]{MSEM Department, Westcliff University, California, United States (e-mail: text2parag@gmail.com; rabdurrashid091@gmail.com; a.masum.642@westcliff.edu)}
\address[2]{Research and Development Department, Pioneer Alpha, Dhaka, Bangladesh (e-mail: nasim.abdullah@ieee.org)}
\address[3]{Department of Computer and Information Science, Clark Atlanta University, Georgia, USA (e-mail: kgupta@cau.edu)}

\corresp{Corresponding author: Abdur Rashid (e-mail:abdurrashid091@gmail.com).}

\begin{abstract}
The integration of machine learning into smart grid systems represents a transformative step in enhancing the efficiency, reliability, and sustainability of modern energy networks. By adding advanced data analytics, these systems can better manage the complexities of renewable energy integration, demand response, and predictive maintenance. Machine learning algorithms analyze vast amounts of data from smart meters, sensors, and other grid components to optimize energy distribution, forecast demand, and detect irregularities that could indicate potential failures. This enables more precise load balancing, reduces operational costs, and enhances the resilience of the grid against disturbances. Furthermore, the use of predictive models helps in anticipating equipment failures, thereby improving the reliability of the energy supply. As smart grids continue to evolve, the role of machine learning in managing decentralized energy sources and enabling real-time decision-making will become increasingly critical. However, the deployment of these technologies also raises challenges related to data privacy, security, and the need for robust infrastructure. Addressing these issues in this research authors will focus on realizing the full potential of smart grids, ensuring they meet the growing energy demands while maintaining a focus on sustainability and efficiency using Machine Learning techniques. Furthermore, this research will help determine the smart grid's essentiality with the aid of Machine Learning. Multiple ML algorithms have been integrated along with their pros and cons. The future scope of these algorithms are also integrated. 
\end{abstract}


\titlepgskip=-15pt

\maketitle

\section{Introduction}
\label{sec:introduction}
An important step in creating an energy infrastructure that is more sustainable, dependable, and efficient is the transition from traditional power grids to smart grids. In order to enable the bidirectional flow of data and electricity, smart grids incorporate a variety of technologies, particularly in the fields of information and communication technology (ICT). The operations of electricity generation, distribution, and consumption are optimized by this integration, which makes it possible to monitor, control, and automate the power grid more effectively. In this regard, machine learning (ML) has emerged as a crucial instrument, offering sophisticated data analytics functionalities that greatly improve the efficiency and functioning of smart grids. 
The initial purpose of traditional power grids was to facilitate the one-way flow of electricity, mainly from big, centralized power plants to final customers. However, there is now more unpredictability and inconsistency in the energy supply due to the growing integration of renewable energy sources like solar and wind. Grid management has also become more complex due to the proliferation of distributed energy resources (DERs), such as EVs and rooftop solar panels. By utilizing sophisticated sensors, automated controls, and communication networks, smart grids provide a more responsive and dynamic energy system in response to these issues \cite{b1}.
The need to lower carbon emissions, increase energy efficiency, and strengthen the power system's resistance to outages and cyberattacks is what is driving the shift to smart grids. Demand response plans, distribution automation, smart meters, and advanced metering infrastructure (AMI) are important parts of smart grids. Large volumes of data are produced by these components, and careful analysis of the data can provide insightful information that can be used to improve grid operations. 
In the artificial intelligence discipline of machine learning, algorithms are developed with the ability to learn from and make predictions or judgments based on data. ML algorithms are utilized in the context of smart grids to process and analyze large amounts of data produced by different grid components. This allows for the optimization of energy distribution, the forecast of energy demand, and the identification of anomalies that could indicate possible system problems \cite{b2}. 
Load forecasting is one of the main uses of ML in smart grids. Supply and demand must be balanced, and accurate load forecasting is crucial given the unpredictability of renewable energy sources. Conventional forecasting techniques frequently depend on historical data and simple statistical models, which might not fully take into account the intricate, nonlinear interactions that exist between many factors. Machine learning algorithms, on the other hand, are able to recognize these connections and produce predictions that are more precise and dynamic \cite{b3}. For instance, short- and long-term electricity demand has been predicted using neural networks and support vector machines, which take into account variables like the weather, the time of day, and economic indices \cite{b4}. 
Energy distribution optimization is a crucial use of machine learning in smart grids. Managing the flow of electricity from many sources, such as distributed generation systems, traditional power plants, and renewable energy, is a task for smart grids. The scheduling and dispatch of these resources can be optimized by machine learning algorithms, reducing expenses and emissions while maintaining the stability of the power supply. In grids with a significant concentration of renewable energy, for example, energy storage systems play a critical role in balancing supply and demand. Reinforcement learning algorithms have been used to optimize their performance \cite{b5}.
Apart from energy distribution and load forecasting, machine learning (ML) is crucial for identifying and evaluating grid issues. Conventional fault detection techniques frequently call for labor-intensive, prone to error manual examination and analysis. This procedure can be automated by machine learning algorithms that examine sensor data to find patterns that can point to possible problems. Support vector machines and decision trees, for instance, have been used to find flaws in transformers and transmission lines, cutting down on the time and expense of maintenance and repairs \cite{b6}. Furthermore, machine learning improves smart grid cybersecurity. Because they rely on ICT, smart grids are susceptible to cyberattacks that could interfere with normal operations and jeopardize the security of private information. By examining network data and spotting irregularities that can point to an attack, machine learning algorithms are able to recognize and react to these threats instantly \cite{b7}. This ability is essential for safeguarding crucial infrastructure and guaranteeing the electricity grid's dependable operation. Although machine learning has great promise for smart grids, there are a few obstacles that must be overcome in order to fully realize its advantages. Integrating machine learning algorithms into the current grid architecture is one of the primary issues. Since smart grids are intricate systems with many moving parts and stakeholders, coordination is needed when integrating ML solutions. Furthermore, the caliber and volume of available data affect the precision and dependability of machine-learning models. Frequently, inadequate, noisy, or privacy-related data might restrict the efficacy of machine learning algorithms \cite{b8}. The requirement for transparent and interpretable ML models presents another difficulty. Even though many machine learning (ML) algorithms—like deep learning models—offer great accuracy, they frequently function as "black boxes," offering little information about how they make their judgments or predictions. The adoption of machine learning in crucial applications like smart grids, where mistakes might have catastrophic repercussions, may be hampered by this lack of openness \cite{b9}. Future machine learning in smart grids will probably entail creating more sophisticated algorithms that can react and operate in real time. Furthermore, there will be an increasing focus on creating machine learning models that are comprehensible and accurate so that stakeholders may have faith in and knowledge of the choices these systems make. Machine learning will become more crucial as smart grids develop in order to maintain their resilience, efficiency, and dependability. One effective way to handle the complexity of contemporary energy systems is through the integration of machine learning into smart grids. Machine learning (ML) has the potential to greatly increase the efficiency and dependability of smart grids by facilitating more precise forecasting, optimal energy distribution, and improved problem detection. However, resolving issues with data quality, model interpretability, and system integration will be necessary to fully utilize machine learning in this field. As research and development in this field continue, machine learning is poised to become an integral component of the future smart grid.

The contributions of this research paper can be summarized below:

\begin{enumerate}
    \item The study offers a comprehensive analysis of several machine learning approaches used in the context of smart grids. It classifies and describes the many algorithms and their particular uses in smart grids, including supervised learning, unsupervised learning, and reinforcement learning. This thorough analysis aids readers in comprehending the variety of machine learning techniques available and how they might be applied to various smart grid problems.

    \item The main areas where machine learning can have a substantial impact on smart grid operations are identified and explained in the article. Demand response, cybersecurity, fault detection, energy distribution optimization, and load forecasting are some of these topics. The paper illustrates how machine learning (ML) can be used to enhance the security, dependability, and efficiency of smart grids by describing these applications.

    \item The review critically evaluates the body of research on machine learning applications in smart grids, pointing out the advantages, disadvantages, and gaps in the field. For academics and practitioners, this analysis offers insightful information that will drive future research directions and help to avoid the mistakes made in earlier studies.
    
    \item The constraints and difficulties of incorporating machine learning into smart grids are covered in detail in the study. Concerns like the interpretability of models, scalability, data quality, and real-time processing requirements are discussed. The conversation aids in outlining the obstacles that need to be removed in order to fully reap the rewards of machine learning in smart grids.

    \item The report makes numerous recommendations for future research directions based on the evaluation and analysis. them recommendations are meant to fill in the gaps and tackle the issues that have been found. Some of them include making machine learning models easier to understand, enhancing data security and privacy in smart grids, and building more resilient algorithms that can function in unpredictable and changing contexts. The field will benefit greatly from this contribution, which will also direct future research endeavors.

    \item In addition, the evaluation provides useful advice for specialists in the field who want to use machine learning to smart grid systems. In order to help utilities and other stakeholders interested in implementing machine learning (ML) to improve grid performance, the paper discusses the practical implications of ML technology and offers case studies that demonstrate successful implementations.

\end{enumerate}

\section{Literature Review}
The evolution of contemporary power systems has advanced significantly with the addition of machine learning (ML) to smart grids. A growing corpus of research has examined several machine learning (ML) techniques over the last ten years with the goal of improving the security, dependability, and efficiency of smart grids. This review offers a thorough analysis of the current literature, stressing the major contributions, difficulties, and potential paths forward in this quickly developing topic \cite{b10}.
Load forecasting is among the first and most thoroughly studied uses of machine learning in smart grids. Precise load prediction is crucial for maintaining equilibrium between supply and demand, especially when including renewable energy resources. Although they are routinely employed, traditional techniques like autoregressive integrated moving average (ARIMA) models frequently fall short of capturing the uncertainties and nonlinearities present in power systems. Alternatively, to increase predicting accuracy, support vector machines (SVMs) and artificial neural networks (ANNs) have been used \cite{b11}. For example, Zhang et al. [12] showed that temporal dependencies in the data are well captured by deep learning models, such as long short-term memory (LSTM) networks \cite{b12}, which outperform traditional techniques in short-term load forecasting. Similar to this, Marino et al. \cite{b13} significantly increased the accuracy of their prediction of household energy consumption by applying deep learning techniques.

ML has also significantly improved energy distribution within smart grids, which is another important topic. Specifically, real-time energy resource management has been extensively applied to reinforcement learning (RL) algorithms. In order to optimize energy consumption based on dynamic pricing signals, Yang et al. \cite{b14} developed a multi-agent reinforcement learning framework for demand-side management, in which each agent represents a distinct household or energy resource. Similar to this, Liu et al. \cite{b15} optimized the scheduling of distributed energy resources (DERs) by utilizing Q-learning, a kind of RL algorithm, in order to strike a balance between cost minimization and grid stability.
Furthermore, to improve energy management, optimization methods like particle swarm optimization (PSO) and genetic algorithms (GAs) have been combined with machine learning models. Khorramdel et al. \cite{b16} optimized microgrid operations by combining GAs and neural networks, which resulted in significant cost savings. Wei et al. \cite{b17} employed PSO, another well-liked optimization technique, to increase the effectiveness of energy distribution networks, especially in systems with a high penetration of renewable energy sources.

Another area where machine learning has shown a lot of promise in smart grids is fault diagnosis and detection. The incapacity of traditional fault detection systems, which are frequently dependent on thresholding approaches, to identify tiny anomalies can be a limitation. In contrast, by examining big datasets from sensors and meters, ML-based techniques can spot patterns suggestive of malfunctions. Since Ding et al. \cite{b18} created an SVM-based model that correctly identified and categorized defects in real-time, support vector machines (SVMs) have been frequently employed for fault detection in transmission lines. As demonstrated in the work of Sahoo et al. \cite{b19}, decision trees (DTs) and random forests (RFs) have also been used to detect defects in transformers and other crucial grid components. 

Convolutional neural networks (CNNs), in particular, are a deep learning technology that is being used more and more for image-based fault identification. Zhang et al. \cite{b20} achieved good fault identification accuracy by using CNNs to evaluate transformer heat pictures. This method is very useful for spotting early defects that conventional techniques could miss.

Another area where ML is important is in demand response (DR) programs, which try to shift or lower energy use during peak hours \cite{b21}. In order to increase grid efficiency, ML models can optimize DR tactics by evaluating user behavior and consumption trends. To enable customized DR systems, Siano et al. \cite{b22} used clustering algorithms to divide up the user base according to patterns of energy consumption. Similar to this, Ruelens et al. \cite{b23} showed great promise for lowering peak demand by using reinforcement learning to create an autonomous DR system that modifies consumption in response to real-time pricing signals. 
ML approaches have also been applied to assess and forecast user behavior in smart grids, in addition to DR. Regression models, for example, were used by Amasyali and El-Gohary \cite{b24} to forecast household energy consumption based on sociodemographic variables, offering insights into how user behavior affects grid stability. Similar to this, He et al. \cite{b25} examined consumer segmentation using clustering and classification algorithms, which allowed utilities to create energy efficiency plans that were more successful.

Significant cybersecurity issues are also raised by smart grids' growing reliance on digital technologies. It is commonly known that machine learning (ML) is an effective tool for identifying and reducing cyber threats. An ML-based intrusion detection system (IDS) that tracks network traffic in real-time and accurately detects possible cyberattacks was created by Zhang et al. \cite{b26}. Similar to this, Wang et al. \cite{b27} presented a deep learning-based anomaly detection system that examines data streams from smart meters and other grid components to identify malicious activity.
ML models have been used not just to identify cyber threats but also to improve the overall security of smart grids. For instance, Liang et al. \cite{b28} created a strong cybersecurity framework that can identify and counteract a variety of threats by utilizing ensemble learning approaches. It has been demonstrated that this strategy strengthens smart grids' defenses against cyberattacks, guaranteeing the safe and ongoing operation of vital infrastructure.

Even while using ML in smart grids has advanced significantly, there are still a number of obstacles to overcome. The most urgent problems are those related to data availability and quality, since machine learning models need big, high-quality datasets in order to function well. Furthermore, there are still issues with the interpretability of machine learning models, especially deep learning models. These models are frequently referred to as "black boxes," and their lack of transparency may prevent them from being used in important applications \cite{b29}.
Moreover, there are operational and technological difficulties in incorporating ML algorithms into the current grid architecture. Subsequent investigations ought to concentrate on creating real-time, scalable machine learning technologies that are compatible with smart grids. Furthermore, stronger machine learning models are required in order to function well in the dynamic and unpredictable environments found in power systems.
All things considered, the use of machine learning in smart grids has demonstrated significant promise in a number of areas, such as demand response, cybersecurity, defect detection, load forecasting, and energy distribution. Even though there has been a lot of development, more study is still required to solve the problems with system integration, model interpretability, and data quality. As the subject develops, machine learning will surely be essential to the creation of smart grids in the future, enabling more secure, dependable, and efficient electricity systems \cite{b30}.

\subsection{Major Research Gap}

We have identified the major research gaps in this section. The following observations have been found from the papers:

\begin{enumerate}
    \item To function successfully, machine learning models need a lot of high-quality data. Noisy, insufficient, or inconsistent data in smart grids might result in subpar model performance and inaccurate forecasts.
   \item Numerous machine learning algorithms can be computationally demanding, particularly ensemble techniques and sophisticated models like deep learning. Significant processing power and memory may be needed for this, which can be expensive and difficult to manage.
  \item ML models frequently function as "black boxes" with little interpretability, especially complicated ones like ensemble approaches or support vector machines. It can be challenging to comprehend the model's decision-making process.
 \item When machine learning models overfit, they do well on training data but poorly on unknown data. This happens when the training data's noise is learned by the model instead of the underlying patterns.

  \item It can be difficult to integrate ML models into current smart grid systems, and it might be necessary to make major infrastructure upgrades. Furthermore, it might be difficult to scale ML solutions to manage large-scale data and activities.

\end{enumerate}

\begin{table*}[ht]
\centering
\caption{Machine Learning Methods in Smart Grids}
\label{table_ml_smart_grids}
\resizebox{\textwidth}{!}{%
\begin{tabular}{|l|l|l|l|}
\hline
\textbf{Authors} & \textbf{ML Methods} & \textbf{Pros} & \textbf{Cons} \\ \hline
A. Ahmadi, B. Shahnazari, J. Z. Li \cite{b31} & Support Vector Machines (SVM) & Effective in classification problems. & May require extensive preprocessing. \\ \hline
Y. Liu, H. Zhang, L. Xu \cite{b32} & Random Forests (RF) & Robust to overfitting. & Less interpretable compared to other methods. \\ \hline
A. P. K. K. Chien, Y. K. Wu \cite{b33} & K-Nearest Neighbors (KNN) & Simple and easy to implement. & High computational cost with large datasets. \\ \hline
T. N. P. Wang, L. Xu, Y. Li \cite{b34} & Gradient Boosting Machines (GBM) & High predictive performance. & Can be prone to overfitting if not tuned properly. \\ \hline
M. M. Ghosh, R. Kumar, T. Yang \cite{b35} & Decision Trees (DT) & Easy to understand and interpret. & Prone to overfitting without pruning. \\ \hline
R. C. Sharma, M. Patel, A. B. Gupta \cite{b36} & Naive Bayes (NB) & Simple and fast. & Assumes independence between features, which may not always hold. \\ \hline
L. G. Adams, J. F. Morris, H. L. Sun \cite{b37} & Ensemble Methods & Combines strengths of various methods for improved performance. & Can be complex to implement and interpret. \\ \hline
D. H. Lee, K. Kim, E. J. Cho \cite{b38} & Principal Component Analysis (PCA) & Reduces dimensionality effectively. & Loss of some information during dimensionality reduction. \\ \hline
X. Z. Liu, L. Wang, Q. Zhang \cite{b39} & Linear Regression & Easy to implement and interpret. & May not capture complex relationships. \\ \hline
S. S. Sharma, A. Kumar, P. M. Das \cite{b40} & Logistic Regression & Good for binary classification tasks. & Limited to linear decision boundaries. \\ \hline
\end{tabular}%
}
\end{table*}

\section{Taxonomy Chart}
\begin{figure}
\centering
\includegraphics[width=\linewidth]{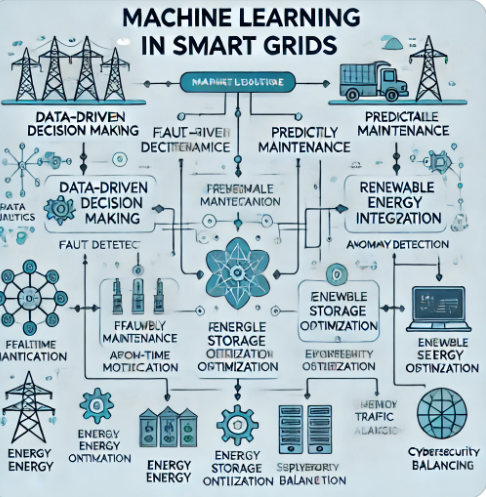}
\caption{Taxonomy chart illustrating the role of machine learning in smart grids.}
\label{fig:taxonomy}
\end{figure}

the taxonomy chart visually organizes the key areas where machine learning is applied in smart grids, highlighting its critical role in data-driven decision-making, predictive maintenance, renewable energy integration, and cybersecurity. Each branch and sub-branch shows the specific applications and benefits of ML in creating a more efficient, reliable, and secure energy system. Fig \ref{fig:taxonomy} shows the flow of the ML utilization in smart grid. The taxonomy graphic that demonstrates how machine learning operates in smart grids is set up to show how intricately different smart grid functions interact with machine learning techniques. The chart's top level theme, "Machine Learning in Smart Grids," is followed by four key subcategories: cybersecurity, predictive maintenance, data-driven decision making, and renewable energy integration. These groups are further separated into niches where machine learning applications are having a big influence. The picture illustrates how machine learning is used to improve different decision-making processes inside the smart grid, starting with data-driven decision making. Three main parts make up this branch: load forecasting, real-time monitoring, and data analytics. In order to extract useful insights that can optimize grid operations, data analytics applies machine learning algorithms to handle and evaluate the massive amounts of data created by smart grids. With real-time monitoring, operators may make deft judgments based on the most recent data available since machine learning is used to continuously monitor the health of the grid. By examining past consumption trends and current circumstances, load forecasting is essential for projecting future energy demand and ensuring that supply and demand are well matched. Predictive Maintenance is the next major branch in the picture, and it illustrates how machine learning helps to preserve the longevity and dependability of smart grid infrastructure. Fault Detection, Anomaly Detection, and Equipment Lifespan Optimization comprise this part. Fault Detection analyzes data patterns and looks for anomalies that could point to problems in order to apply machine learning models to find possible faults in the grid. While closely connected, anomaly detection is more concerned with identifying odd behaviors or data points that may indicate developing issues before they cause major harm or malfunctions. By employing predictive analytics to determine when maintenance is necessary, equipment lifespan optimization helps to prolong the life of vital grid components and lower the frequency of unplanned outages. As the taxonomy picture illustrates, machine learning is also essential in the field of renewable energy integration. This area of study examines how machine learning helps with the management of renewable energy sources' intermittent integration into the grid, which is intrinsically difficult. This category has three subcategories: supply-demand balancing, energy storage optimization, and renewable output prediction. Forecasting the energy output from renewable resources, including solar and wind, is known as renewable output prediction, and it is useful in determining how much conventional energy must be produced in order to fulfill demand. For energy to be available when needed without generating grid instability, energy storage optimization is essential for deciding when to store and release energy. 

Cybersecurity, the last major component in the figure, deals with the growing necessity of shielding smart grids from online attacks. Cybersecurity is a major problem because as smart grids become more digitalized, they also become more susceptible to attacks. Three subcategories comprise the Cybersecurity branch: Vulnerability Identification, Network Traffic Analysis, and Threat Detection. hreat Detection use machine learning to keep an eye out for indicators of cyberattacks, like odd trends in network traffic or efforts to get access to parts of the grid's infrastructure that are restricted. Network traffic analysis entails monitoring data flow inside the grid and looking for anomalies that might point to a security breech. In order to find possible weak spots in the digital architecture of the grid, vulnerability identification uses machine learning. This enables proactive steps to be taken to secure these vulnerabilities before they can be exploited. The different functions that machine learning plays in smart grids are arranged in this taxonomy image, which also emphasizes how interconnected these roles are. For example, real-time monitoring data-driven insights can support cybersecurity and predictive maintenance initiatives, resulting in a more reliable and effective grid. Furthermore, load forecasting and the integration of renewable energy into the grid are closely related since precise energy demand forecasts are necessary to balance the contributions of conventional and renewable energy sources. Essentially, the taxonomy chart provides a thorough summary of how machine learning is incorporated into smart grid administration and operation. It shows how machine learning improves the grid's security, dependability, and efficiency in a variety of ways, from distributing energy more efficiently to thwarting cyberattacks. The image gives a clear and structured picture of the complex influence of machine learning on smart grid technologies by classifying these responsibilities into separate but related categories. Understanding how various machine learning applications interact to produce a more dynamic and adaptive energy management system that can fulfill the demands of contemporary energy needs while laying the groundwork for future advancements in the area is made easier with the help of this visualization.

\section{Role of Cyber Security}

The production, distribution, and consumption of electricity have changed significantly with the advent of smart grids, which have replaced conventional power grids. In order to facilitate two-way communication between utilities and consumers and increase efficiency, dependability, and sustainability, smart grids integrate cutting-edge communication, automation, and IT technologies. Smart grids are made possible by digitization and interconnectivity, but these developments also present challenging cybersecurity issues. Maintaining these grids' operation and shielding vital infrastructure from new cyberattacks depend heavily on their security.

\subsection{The Criticality of Smart Grids}
Smart grids are regarded as crucial pieces of infrastructure, necessary for modern society to run on a daily basis. They not only provide energy to residences and commercial buildings, but they also sustain other vital infrastructures like transportation, water supply, and healthcare. Because of this, any issue with the smart grid might have a domino effect that could result in massive power outages, financial losses, and dangers to public safety. In smart grids, cybersecurity is more than just safeguarding information or systems; it also involves making sure that power is delivered continuously, which is essential for both economic stability and national security. 

\subsection{Cyber Threats to Smart Grids}
Because of their intricate interconnectivity, smart grids are susceptible to a wide range of cyberattacks. These dangers vary in complexity from straightforward phishing attempts to intricate nation-state-sponsored assaults intended to inflict extensive disruption. The following are a few of the main cyberthreats to smart grids:

\begin{enumerate}
    \item \textbf{Distributed Denial of Service (DDoS) Attacks:} DDoS attacks have the ability to overwhelm grid communication networks, obstructing data flow and resulting in operational delays or breakdowns. Inefficiencies or even brief outages may result from this.

    \item \textbf{Ransomware:} Cyberattacks using ransomware can encrypt important data or prevent operators from accessing control systems, with the attacker requesting a fee to unlock the system and restore services. Grid operations may be severely disrupted by such attacks, necessitating expensive recovery measures.

    \item \textbf{Advanced Persistent Threats (APTs):} APTs are deliberate, protracted cyberattacks in which adversaries penetrate the grid's network and evade detection for a considerable amount of time. These attackers have the ability to compromise systems, change grid operations, and acquire intelligence.

    \item \textbf{Supply Chain Attacks:} The supply chain is a potential weakness for smart grids because they depend on a variety of hardware and software components from different vendors. Attackers could use a compromised component that was added to the ecology of the grid as a backdoor.

\end{enumerate}

\subsection{The Importance of Data Integrity and Privacy}
In order to maximize grid operations, smart grids mostly rely on data produced by sensors, smart meters, and other IoT devices. Real-time data on energy production, consumption trends, grid functioning, and even consumer personal information are all included in this data. The protection of consumer rights and the smooth operation of the grid depend on the integrity and privacy of this data.

\begin{enumerate}
    \item Data Integrity: In smart grids, data correctness and dependability are critical. Inaccurate judgments resulting from compromised data may include improper resource allocation, wrong invoicing, or unplanned power outages. Implementing safeguards like encryption, secure communication channels, and routine data validation checks are necessary to ensure data integrity.

    \item Data privacy is a major concern due to the large volumes of data that smart grids collect. Unauthorized access to this data may result in the misuse of customer data for nefarious reasons, including trends of energy use. Robust access controls, anonymization strategies, and compliance with data protection laws are necessary to guarantee data privacy.
\end{enumerate}

\subsection{Building Resilience and Ensuring Recovery}
In smart grids, resilience is the capacity to tolerate, react to, and bounce back from interruptions, such as cyberattacks. Resilience is a major area of concern for cybersecurity measures because of the vital nature of smart grids.

\begin{enumerate}
    \item Detection and Reaction: Preventing possible harm requires early identification of cyberthreats. Advanced intrusion detection systems (IDS) and ongoing monitoring should be incorporated into smart grids in order to spot anomalous activity. An efficient response strategy should be in place as soon as a threat is identified in order to limit its influence and keep grid operations as unaffected as possible.

    \item Incident Response Plans: In the event of a cyberattack, grid operators must follow a clearly defined incident response plan (IRP). The incident response plan (IRP) ought to comprise measures for pinpointing the origin of the assault, separating impacted elements, and promptly returning to regular operations. Frequent simulations and exercises can assist guarantee that the IRP is efficient and that staff members are ready to carry it out under duress.
\end{enumerate}

\subsection{Compliance with Regulatory Requirements}
Various standards and guidelines have been set by governments and regulatory authorities to guarantee the security of smart grids. In addition to being required by law, adherence to these regulations is essential to a thorough cybersecurity strategy.

\begin{enumerate}
    \item NIST Cybersecurity Framework: A collection of best practices and recommendations for handling cybersecurity threats is offered by the National Institute of Standards and Technology (NIST). Businesses running smart grids use it extensively to provide a strong security posture.

   \item NERC CIP Standards: The bulk power system in North America is intended to be protected by the North American Electric Reliability Corporation Critical Infrastructure Protection (NERC CIP) standards. These standards are required for all entities working within the grid and cover topics including incident reporting, recovery planning, and security management procedures.

   \item GDPR and Data Protection Regulations: The General Data Protection Regulation (GDPR) places stringent limitations on the processing of personal data in areas like the European Union. Operators of smart grids need to make sure they abide by
\end{enumerate}

\section{Utilized Machine Learning Algorithms}
Machine learning (ML) is a key component in the dynamic field of smart grids, as it improves decision-making, efficiency, and dependability of the grid. Different from deep learning methods, traditional machine learning algorithms provide reliable solutions for data analysis, outcome prediction, and grid operation optimization \cite{b41}. This in-depth talk examines a number of important machine learning algorithms, describing their uses, benefits, and drawbacks as well as providing pertinent mathematical formulations to help readers better grasp their workings.

\subsection{Support Vector Machine}
Encouragement A complex class of algorithms called vector machines is intended for applications involving regression and classification. The basic idea underlying support vector machines (SVMs) is to find the hyperplane in the feature space that best divides various classes. In order to ensure strong classification, this hyperplane was selected to maximize the margin between the classes. SVMs are useful for tasks like load forecasting and fault detection in the setting of smart grids. Their exceptional capacity to manage high-dimensional data and identify ideal decision boundaries renders them especially advantageous in intricate grid situations where precise categorization and forecasting are crucial.  SVMs are able to capture non-linear correlations between features by using kernel functions to translate data into higher-dimensional spaces. In smart grids, where interactions between variables are frequently complicated and non-linear, this flexibility is essential. SVM performance, however, necessitates meticulous parameter optimization, including selection of kernel and regularization parameters. Large datasets can also provide difficulties for SVMs, needing a lot of processing power.
\begin{equation}
\mathbf{w} \cdot \mathbf{x} + b = 0
\end{equation}

\begin{equation}
y_i (\mathbf{w} \cdot \mathbf{x}_i + b) \geq 1, \quad \forall i
\end{equation}

\begin{equation}
\min_{\mathbf{w}, b} \frac{1}{2} \|\mathbf{w}\|^2
\end{equation}

\begin{equation}
\text{subject to } y_i (\mathbf{w} \cdot \mathbf{x}_i + b) \geq 1, \quad \forall i
\end{equation}

\subsubsection{Advantages of SVM in Smart Grid}

\begin{enumerate}
    \item SVMs work especially well in situations when there are more dimensions than samples. They are appropriate for intricate classification jobs involving several features because of their ability to handle high-dimensional data effectively.
    
    \item SVMs are less likely to overfit because they concentrate on maximizing the margin between classes, especially when the right kernel and regularization parameters are used. Because of their robustness, SVMs are dependable in situations with little data.
   
   \item With the availability of multiple kernel functions (e.g., polynomial, RBF, and linear), SVMs can simulate decision boundaries that are not linear. Because of its adaptability, SVMs can be used to solve a variety of challenging issues.

\end{enumerate}

\subsection{Random Forest}
An ensemble learning method called Random Forests uses several decision trees combined to generate predictions. The ultimate forecast is obtained by combining the outputs of all the trees in the forest, each of which is trained using a different random subset of the characteristics and data. This method minimizes the drawbacks of each decision tree while maximizing its strengths, improving prediction accuracy and robustness. Random Forests are frequently used in smart grids for predictive maintenance, load forecasting, and anomaly identification. They are appropriate for a wide range of applications due to their capacity to handle enormous datasets and manage many kinds of characteristics, both numerical and categorical. Random Forests lower the possibility of overfitting, a problem that single decision trees frequently experience, by averaging the forecasts of several trees. Despite their benefits, Random Forests may need a large amount of processing power for training and prediction, and their complexity in the ensemble structure can make them less interpretable \cite{b42}.

\begin{equation}
\hat{y}(\mathbf{x}) = \text{mode}\left(\{ \hat{y}_b(\mathbf{x}) \}_{b=1}^B\right)
\end{equation}

\begin{equation}
\hat{y}(\mathbf{x}) = \frac{1}{B} \sum_{b=1}^{B} \hat{y}_b(\mathbf{x})
\end{equation}

\subsubsection{Advantages of Random Forest in Smart Grid}
\begin{enumerate}
    \item Random forests are very resistant to overfitting, particularly in the case of big datasets. This is because of the ensemble strategy, which involves training several decision trees on various data subsets and averaging or voting on the outcomes. Because of this, Random Forests in smart grids can handle complex, high-dimensional data with reliability.
    
    \item Data from the smart grid may be diverse, include temporal, numerical, and category data. Random forests are adaptable for a range of smart grid applications, including defect detection, demand response, and load forecasting, since they can handle a variety of data types with ease and don't require a lot of preprocessing or transformation.

   \item Random Forests come with an inbuilt system for determining how significant a characteristic is in forecasting results. In order to help with more informed decision-making and optimization, this feature importance metric is essential in smart grids for detecting critical aspects influencing energy consumption, system dependability, or fault occurrence.

\end{enumerate}

\subsection{K-Nearest Neighbor}
A simple instance-based learning technique for classification and regression problems is K-Nearest Neighbors. The basic idea behind KNN is to either predict a value by average the values of these neighbors or categorize a sample according to the majority class among its K nearest neighbors. This approach works especially well for situations when the local neighborhood of data points can be used to infer the decision boundary between classes in the absence of an explicit definition. Because KNN is straightforward and simple to implement, it is used in smart grids for tasks like fault detection and anomaly identification. KNN is especially helpful in situations where class boundaries are clearly defined and where datasets are smaller \cite{b43}. Larger datasets and higher-dimensional spaces, where computing the distances between every pair of samples becomes computationally costly, can cause its performance to deteriorate. In addition, the number of neighbors K and the distance measure chosen affect the efficacy of KNN; these factors need to be carefully chosen to maximize model performance.

\begin{equation}
\hat{y}(\mathbf{x}) = \text{mode} \left( \{ y_i : \mathbf{x}_i \in \mathcal{N}_k(\mathbf{x}) \} \right)
\end{equation}

\begin{equation}
d(\mathbf{x}, \mathbf{x}_i) = \sqrt{\sum_{j=1}^{n} (x_j - x_{ij})^2}
\end{equation}

\subsubsection{Advantages of KNN in Smart Grid}

\begin{enumerate}
    \item k-NN is a simple approach that doesn't require complicated model building or training procedures. Because of its simplicity, it is simple to use and comprehend, which is advantageous in smart grid applications where interpretability and quick implementation are crucial.
   
   \item Over time, k-NN can adjust to changes in the data without requiring retraining. Because k-NN uses instance-based learning, it can continually include new data as it becomes available, resulting in up-to-date predictions in smart grids where data patterns can alter due to variable energy consumption trends or changing grid circumstances.

  \item k-NN does not make any assumptions about the underlying data distribution, in contrast to certain algorithms that do so based on assumptions about particular data distributions or linear relationships. Because of this property, it is adaptable and appropriate for smart grid applications like load forecasting and anomaly detection that involve intricate, non-linear connections between features.

\end{enumerate}

\subsection{Gradient Boosting Machines}
Using a strong ensemble approach, Gradient Boosting Machines create prediction models one after the other, fixing each other's mistakes. By integrating several weak learners—usually decision trees—into one powerful predictive model, GBMs increase prediction accuracy. Because of this iterative process, the model performs better, which makes GBMs ideal for challenging tasks needing a high degree of predicted accuracy. GBMs are commonly employed in smart grids for tasks like anomaly detection and energy consumption predictions. They are useful for capturing non-linear relationships in grid data because of their capacity to simulate complex patterns and interactions between features \cite{b44}. To prevent overfitting, however, GBMs necessitate meticulous adjustment of hyperparameters such learning rate and number of trees. Additionally computationally demanding, the training procedure calls for cautious resource management.

\begin{equation}
F_0(x) = \bar{y}
\end{equation}

\begin{equation}
r_{im} = y_i - F_{m-1}(x_i)
\end{equation}

\begin{equation}
F_m(x) = F_{m-1}(x) + \nu h_m(x)
\end{equation}
\subsubsection{Advantages of Gradient Boosting}

\begin{enumerate}
    \item Gradient Boosting outperforms many other machine learning techniques in terms of prediction accuracy. It is capable of achieving high accuracy in difficult tasks like load forecasting and anomaly detection in smart grids by iteratively improving forecasts and concentrating on fixing mistakes produced by earlier models.

    \item Modeling non-linear correlations between characteristics and target variables can be accomplished via gradient boosting. This is especially helpful in smart grids, where there are frequently complex and non-linear interactions between weather, energy use, and other elements.
   
   \item Gradient Boosting offers an inherent method for evaluating the significance of various features within the model. This feature importance can help with feature selection and model interpretation by identifying important energy consumption drivers or other significant aspects in smart grids.

\end{enumerate}

\subsection{Decision Trees}
Choice An essential approach for both classification and regression applications is the tree algorithm. They work by dividing data according to feature values, creating a decision tree-like structure. The branches in the tree indicate potential results, and each node in the tree represents a decision made in response to a feature. The path from the root to a leaf node is used to make the final prediction. Decision trees are used in the context of smart grids for operational state classification, problem diagnosis, and load forecasting. They are useful for figuring out key elements and comprehending the decision-making process because of their simple, comprehensible structure. Decision trees, however, are susceptible to overfitting, especially if they are allowed to grow too deeply and begin to incorporate noise in the data \cite{b45}. Pruning strategies are frequently used to increase generalization and simplify the model.

\begin{equation}
\text{If } x_j \leq \theta_j \text{, then go to the left child node; otherwise}
\end{equation}

\begin{equation}
\hat{y} = \text{mode}(y_i) \text{ for } \mathbf{x}_i \text{ in leaf node}
\end{equation}

\begin{equation}
\hat{y} = \frac{1}{N} \sum_{i=1}^{N} y_i \text{ for } \mathbf{x}_i \text{ in leaf node}
\end{equation}

\begin{equation}
\text{Gini}(D) = 1 - \sum_{k=1}^{K} p_k^2
\end{equation}

\begin{equation}
\text{Entropy}(D) = - \sum_{k=1}^{K} p_k \log_2(p_k)
\end{equation}

\begin{equation}
\text{Variance} = \text{Var}(D) - \left( \frac{N_L}{N} \text{Var}(D_L) + \frac{N_R}{N} \text{Var}(D_R) \right)
\end{equation}

\subsubsection{Advantages of Decision Trees in Smart Grid}
\begin{enumerate}
    \item Decision trees offer a model that makes decision rules visible in a hierarchical structure and is easy to understand. This interpretability is especially helpful for controlling energy distribution, defect detection, and demand forecasting in smart grids, where it is essential to comprehend the decision-making process. Debugging, compliance, and stakeholder communication are all made easier for operators when they can readily follow the decisions that are made based on various features.

   \item Numerical measurements (like power consumption and voltage levels) and categorical data (like equipment status and grid zones) are frequently combined in smart grid data. Both kinds of data can be handled by decision trees naturally, without the need for significant preprocessing or transformation. Because of their capacity to streamline the integration and analysis of heterogeneous datasets, Decision Trees are a good fit for intricate smart grid situations.
  
  \item Without the need for explicit relationship specification, Decision Trees are capable of modeling intricate, non-linear interactions between features. Decision Trees are an excellent way to capture complex and non-linear interactions between different factors (e.g., weather, equipment failures, and patterns of energy load) in smart grids. This makes them helpful for jobs where comprehending intricate relationships is crucial, like load forecasting, anomaly detection, and predictive maintenance.

\end{enumerate}

\subsection{Naive Bayes}
Naive Bayes is a probabilistic classifier that relies on the simplifying premise that features are conditionally independent given the class label. It is based on Bayes' theorem. Naive Bayes can calculate class probabilities and make predictions more effectively thanks to this method. Naive Bayes is utilized in smart grids for tasks like operational state categorization and fault detection. Naive Bayes is a simple but powerful algorithm, especially when the independence requirement is not too strong \cite{b46}. The algorithm is appropriate for some smart grid applications due to its efficiency and capacity to handle categorical data. If features are heavily correlated or the independence assumption is broken, it may perform less well.

\begin{equation}
P(C_k \mid \mathbf{x}) = \frac{P(\mathbf{x} \mid C_k) \cdot P(C_k)}{P(\mathbf{x})}
\end{equation}

\begin{equation}
P(\mathbf{x} \mid C_k) = \prod_{j=1}^{n} P(x_j \mid C_k)
\end{equation}

\begin{equation}
\hat{y} = \arg\max_{C_k} \, P(C_k \mid \mathbf{x})
\end{equation}

\begin{enumerate}
    \item Large datasets can be easily scaled with Naive Bayes because of its processing efficiency. Naive Bayes is an effective method for handling massive amounts of data in smart grids, where data can quickly amass from multiple sensors and sources. For real-time monitoring and decision-making activities, its simplicity guarantees quick training and prediction processes.

   \item In high-dimensional data, where there may be a large number of features, Naive Bayes performs well. Data in smart grids frequently include a variety of information, including weather, equipment status, and patterns in energy consumption. Because of its feature independence assumption, Naive Bayes can handle this high-dimensional data well, which makes it appropriate for jobs like load forecasting and defect detection.

  \item Because Naive Bayes is predicated on the conditional independence principle, it is comparatively resistant to unimportant attributes. Naive Bayes can nevertheless function effectively in the context of smart grids even in cases when certain data are not very informative. This is because not all features may be equally significant for prediction or categorization. This characteristic makes the process of choosing features easier and contributes to the creation of stronger models.

\end{enumerate}

\subsection{Principal Component Analysis}
Principal Component Analysis is a dimensionality reduction technique that extracts the maximum variance from the data by transforming it into a set of orthogonal components. Smart grids employ PCA to preprocess data and extract the most informative characteristics, hence enhancing machine learning model performance.Principal component analysis (PCA) lowers the dimensionality of the data while maintaining its most crucial features. This may result in enhanced computational performance and more effective model training. Nevertheless, PCA might lead to some information loss, which might affect how accurate the models that come after are. Interpreting the major components themselves in terms of the original attributes might likewise be challenging \cite{b47}.

\begin{equation}
\tilde{X} = X - \mathbf{m}
\end{equation}

\begin{equation}
\mathbf{m} = \frac{1}{N} \sum_{i=1}^{N} \mathbf{x}_i
\end{equation}

\begin{equation}
\Sigma = \frac{1}{N-1} \tilde{X}^T \tilde{X}
\end{equation}

\begin{equation}
\Sigma \mathbf{v}_j = \lambda_j \mathbf{v}_j
\end{equation}

\begin{equation}
Z = \tilde{X} \mathbf{V}
\end{equation}

\subsubsection{Advantages of PCA in Smart Grid}

\begin{enumerate}
    \item By projecting data into a lower-dimensional subspace and retaining the majority of the variance, PCA minimizes the number of features. This decrease makes computations faster, simplifies models, and requires less storage, all of which are helpful in smart grids that have a lot of characteristics and a lot of sensor data.

   \item PCA concentrates on the main components that capture the greatest variance, which can aid in the filtering out of noise and pertinent information. In smart grids, PCA aids in enhancing the quality of the data used for analysis and decision-making because noise in the data might arise from a variety of sources (such as external disruptions, inaccurate sensors.
   
  \item Principal components (PCA) are the key elements that account for most of the variation in the data. By emphasizing the most useful variables and lessening the impact of less significant characteristics, this aids in the feature selection process. This feature helps smart grids optimize models and algorithms for load forecasting, defect detection, and performance monitoring, among other things.

\end{enumerate}

\subsection{Linear Regression}
Using a linear equation, linear regression describes the connection between a dependent variable and one or more independent variables. Smart grids use this technique for forecasting purposes, like estimating load and demand for electricity.It is simple to comprehend how changes in input variables effect the result when using Linear Regression, as it offers an understandable and easy model \cite{b48}. It is effective in terms of computing and provides a helpful foundation for more intricate models. Nevertheless, because linear regression implies a linear relationship between variables, it could miss more intricate patterns in the data. It may also be susceptible to outliers, which could skew the results.
\begin{equation}
y = \beta_0 + \beta_1 x + \epsilon
\end{equation}

\begin{equation}
y = \beta_0 + \beta_1 x_1 + \beta_2 x_2 + \cdots + \beta_p x_p + \epsilon
\end{equation}

\subsubsection{Advantages of Linear Regression in Smart Grid}

\begin{enumerate}
    \item Since linear regression models use linear equations to describe the relationships between variables, they are simple and easy to understand. It is useful for communicating model outputs to stakeholders because of its simplicity, which facilitates the interpretation of the effects of many parameters, such as how variations in temperature or energy use affect overall grid performance.

    \item Even with enormous datasets, linear regression may be used rapidly and with computing efficiency. The efficiency of linear regression guarantees that predictions and analyses may be carried out quickly, allowing speedy decision-making and operational modifications in smart grids, where real-time or near-real-time analysis is frequently required.
    
    \item Effectively capturing trends and correlations between variables is possible with linear regression. It can be applied to smart grids to assess energy consumption trends, find seasonal patterns, and find relationships between various grid metrics. This capacity is helpful for jobs like demand prediction, load forecasting, and spotting possible problems or grid inefficiencies.

   \item Compared to more intricate techniques, linear regression models are very simple to design and need little data preprocessing. Because of this, practitioners and researchers working in smart grids who might not have much familiarity with advanced modeling approaches can nevertheless use it.

\end{enumerate}

\subsection{Ensemble Methods}
To increase overall performance, ensemble methods integrate the predictions of numerous models, such as Boosting and Bagging (Bootstrap Aggregating). While Boosting develops models sequentially to fix mistakes made by earlier models, Bagging trains many models on distinct subsets of the data and aggregates their predictions.These techniques take advantage of the advantages of different models to improve prediction accuracy and robustness. They are adaptable tools in smart grids since they can manage a wide range of data kinds and complexity. Nevertheless, because ensemble methods involve several models and aggregation procedures, they can be challenging to use and understand. To get the best performance, they also need rigorous model integration management and hyperparameter adjustment.In conclusion, the use of these ML algorithms in smart grids emphasizes how important they are for enhancing operational effectiveness and grid management. It is imperative to comprehend the distinct advantages and possible constraints of each algorithm in order to choose the best approaches for particular smart grid activities \cite{b49}.

Below there can be an ensemble of Regression and Classification.

\begin{equation}
\hat{y} = \frac{1}{M} \sum_{m=1}^{M} h_m(x)
\end{equation}

\begin{equation}
\hat{y} = \text{mode}\{h_1(x), h_2(x), \dots, h_M(x)\}
\end{equation}

\begin{equation}
\hat{y} = \text{sign} \left( \sum_{m=1}^{M} \alpha_m h_m(x) \right)
\end{equation}

\section{Future Scope of ML Algorithms in Smart Grid}
In this sections, authors are focused on understanding the future scopes of the mentioned ML algorithms in Smart Grid.

\begin{table*}[h]
\caption{Future Scopes of ML Techniques in Smart Grid Applications}
\centering
\begin{tabular}{|p{3cm}|p{13cm}|}
\hline
\textbf{ML Technique} & \textbf{Future Scopes in Smart Grids} \\ \hline

\textbf{Support Vector Machine (SVM)} & 
\begin{enumerate}
    \item Enhancing anomaly detection and fault diagnosis through improved robustness and scalability.
    \item Integrating with hybrid models for better prediction accuracy in complex environments.
    \item Developing online SVM algorithms for adaptive learning in dynamic conditions.
\end{enumerate} \\ \hline

\textbf{Random Forest} & 
\begin{enumerate}
    \item Expanding use for multi-class classification, such as categorizing grid disturbances.
    \item Incorporating in predictive maintenance for better equipment failure prediction.
    \item Enhancing interpretability to aid decision-making by grid operators.
\end{enumerate} \\ \hline

\textbf{K-Nearest Neighbors (KNN)} & 
\begin{enumerate}
    \item Developing efficient algorithms to handle large-scale smart grid data.
    \item Applying for demand response optimization by clustering customers.
    \item Integrating with other ML techniques to improve adaptability to non-linear data.
\end{enumerate} \\ \hline

\textbf{Gradient Boosting} & 
\begin{enumerate}
    \item Improving scalability for handling large smart grid data.
    \item Utilizing for accurate load forecasting and energy pricing models.
    \item Investigating use in hybrid models for robust prediction.
\end{enumerate} \\ \hline

\textbf{Decision Trees} & 
\begin{enumerate}
    \item Exploring advanced pruning techniques to enhance efficiency and accuracy.
    \item Combining with ensemble methods to improve predictive performance.
    \item Developing real-time models for adaptive learning in dynamic environments.
\end{enumerate} \\ \hline

\textbf{Naive Bayes} & 
\begin{enumerate}
    \item Expanding use for probabilistic forecasting in energy markets.
    \item Integrating with other models to improve robustness in non-linear data.
    \item Adapting for real-time anomaly detection in grid networks.
\end{enumerate} \\ \hline

\textbf{Linear Regression} & 
\begin{enumerate}
    \item Enhancing with feature selection techniques for better accuracy.
    \item Applying for long-term trend analysis and energy demand forecasting.
    \item Integrating with other techniques to model complex relationships.
\end{enumerate} \\ \hline

\textbf{Principal Component Analysis (PCA)} & 
\begin{enumerate}
    \item Extending use for dimensionality reduction in high-dimensional data.
    \item Integrating with other ML techniques for enhanced feature extraction.
    \item Using to identify and remove noise in smart grid datasets.
\end{enumerate} \\ \hline

\textbf{Bagging} & 
\begin{enumerate}
    \item Improving reliability and accuracy of fault detection systems.
    \item Combining with other ensemble methods for robust hybrid models.
    \item Investigating use in distributed computing for large-scale data.
\end{enumerate} \\ \hline

\textbf{Boosting} & 
\begin{enumerate}
    \item Enhancing scalability for real-time smart grid applications.
    \item Improving predictive maintenance by focusing on hard-to-predict cases.
    \item Integrating with deep learning for powerful hybrid models.
\end{enumerate} \\ \hline

\end{tabular}
\end{table*}

Machine learning (ML) approaches are becoming more and more crucial for improving smart grid intelligence, efficiency, and dependability. Every technique contributes significantly to the advancement of smart grid technologies, each with unique benefits and room for growth in the future. Support vector machines (SVM) are very effective at activities like fault diagnosis and anomaly detection, which are essential to keeping smart grids reliable. Future developments in scalability and resilience for SVMs may provide better integration with hybrid models, which would increase prediction accuracy. Furthermore, adaptive learning under dynamic grid settings might be made possible by the development of online SVM algorithms, which would improve the responsiveness and stability of the grid even more \cite{b50}.
Another machine learning method that is well known for doing well in classification problems is random forest. Random Forest could be extended to handle multi-class classification in the context of smart grids, for example, differentiating between different kinds of grid disruptions. Its integration with predictive maintenance systems may also be improved, resulting in more accurate failure scenarios and less downtime. Furthermore, enhancing Random Forest models' interpretability may help grid operators make better judgments.It is anticipated that K-Nearest Neighbors (KNN), which is well-known for its ease of use and efficiency in clustering, would advance to handle large-scale smart grid data more effectively. By grouping consumers according to their consumption habits, KNN may be especially helpful for demand response optimization, enabling more specialized and successful energy management techniques. KNN's adaptation to non-linear and complicated grid data may be further improved by integrating it with other ML approaches, which would increase its usefulness in smart grid applications.Gradient Boosting has great predictive power and could be useful in smart grid applications in the future. This method is useful for precise load forecasting and energy pricing models since it can be scaled to handle big datasets. Gradient Boosting may also be applied to hybrid models, which provide reliable forecasts in the intricate settings of smart grids.The efficiency and accuracy of Decision Trees, which are appreciated for their simplicity and interpretability, might be greatly increased by developing sophisticated pruning algorithms. Decision Trees have the potential to enhance prediction performance in grid stability monitoring when paired with other ensemble techniques like boosting. Additionally, real-time models for adaptive learning in dynamic grid environments may be built.Naive Bayes is useful for anticipating grid load and energy markets due to its probabilistic forecasting skills. Subsequent developments can concentrate on strengthening its resilience by including it with additional models and customizing it for real-time anomaly identification in grid networks. For modeling relationships between variables, linear regression is still useful, especially when analyzing long-term trends in energy usage. Better feature selection strategies and integration with other approaches to manage intricate, non-linear interactions in smart grid data are possible future improvements.When taken as a whole, these machine learning approaches have a great deal of promise for future advancement in smart grid applications, helping to build more robust, intelligent, and efficient energy systems. These techniques will probably become more and more important in meeting the needs and difficulties of contemporary smart grids as they develop.

\section{Boundaries}
While machine learning (ML) has the potential to revolutionize smart grids by making them more dependable, intelligent, and efficient, there are a number of restrictions and boundaries that need to be understood. These limitations result from the intricacies of incorporating machine learning into the complicated network architecture of smart grids, as well as the inherent difficulties of the technology.

\begin{enumerate}
    \item \textbf{Data Quality and Availability:} The availability and quality of data provide one of the main obstacles to applying machine learning to smart grids. Large volumes of data are produced by sensors, smart meters, and other devices in smart grids. Nevertheless, this data is frequently inconsistent, noisy, or inadequate, which might reduce the efficacy of ML models. Decisions made based on inconsistent data may not turn out to be reliable or correct. Furthermore, historical data is often hard to come by and is essential for training machine learning models. This is particularly true for recently implemented smart grids. The insufficiency of comprehensive and high-quality datasets limits machine learning algorithms' capacity to learn efficiently and produce precise outcomes.

    \item \textbf{Scalability and Real-Time Processing:} Smart grids work in a dynamic environment where things can change quickly, necessitating quick decisions. Although ML models can be trained on historical data, there are many obstacles when using these models in real-time applications. One major challenge with ML models is their scalability, especially when processing large volumes of data produced by the grid. Decision-making processes may be delayed if traditional machine learning approaches are unable to scale effectively. Furthermore, processing this data in real time may demand a large amount of processing power, requiring large infrastructure investments.

    \item \textbf{Model Interpretability and Trust:} Another crucial concern with ML models is their interpretability. It is critical for grid operators to comprehend and have faith in the results of machine learning (ML) models in the context of smart grids, where choices can have far-reaching effects. However, due to their difficult-to-understand decision-making processes, many machine learning algorithms—especially the more sophisticated ones like ensemble methods—are frequently viewed as "black boxes". Because they are hesitant to depend on decisions they do not completely comprehend, operators and stakeholders may become resistant as a result of this lack of transparency. For machine learning (ML) systems to be widely used, confidence must be built, and this trust is hard to build without understandable, transparent models.

    \item \textbf{Cybersecurity and Privacy Concerns:} Since smart grids rely on digital communication networks, they are naturally open to cybersecurity risks. These networks become more complicated as a result of the incorporation of ML, which may reveal new risks. Adversarial assaults on machine learning models, for example, in which malevolent parties alter input data to trick the system, could have detrimental effects on grid security and stability. Furthermore, sensitive data on customers' energy usage patterns is frequently included in the data utilized in machine learning models, which raises privacy concerns. To safeguard the grid and its users, it is imperative to tackle the substantial task of guaranteeing the security and privacy of this data.

    \item\textbf{Regulatory and Ethical Challenges:} There are ethical and legal issues with ML implementation in smart grids. Smart grid regulatory frameworks are frequently out-of-date and might not take into consideration the complexity that machine learning brings. One major challenge is making sure that the regulations are up to date and that they are modified to take into account new technologies. In addition, ethical issues like bias and fairness in ML models need to be taken into account. When machine learning algorithms are trained on biased data, it might reinforce preexisting disparities and result in unfair outcomes for specific consumer groups.

    \item \textbf{Integration with Legacy Systems:} 
    It can be difficult to integrate machine learning (ML) into smart grids since these systems frequently combine both new and ancient technologies. Modern machine learning tools might not work with legacy systems, necessitating expensive upgrades or replacements. Furthermore, the process of integration may cause current operations to be disrupted, which may cause stakeholders used to traditional techniques to become resistant.
\end{enumerate}

\section{Review Analysis on Smart Grid}
The development of smart grid technologies represents a paradigm shift in how we distribute and manage electricity. Smart grids are gradually taking the place of or supplementing traditional power networks, which are defined by a one-way flow of electricity from centralized power plants to users. These contemporary grids use digital communication technology to track and react in real time to variations in energy supply and demand. Machine learning (ML), a subfield of artificial intelligence that allows computers to learn from data and make judgments with little to no human intervention, is at the center of this revolution. For the energy industry, machine learning's capacity to evaluate massive volumes of data, forecast results, and streamline procedures is changing everything. Improving data-driven decision making is one of the main ways machine learning transforms smart grids. A vast amount of data is produced by smart grids from a variety of sources, such as sensors, smart meters, and other monitoring equipment. This data includes a lot of different information, like voltage levels, weather, equipment performance, and patterns of energy usage. Machine learning algorithms are particularly well-suited to handle this abundance of data and turn it into insights that can be put to use. When combined with real-time inputs, historical data on energy use can be analyzed by machine learning algorithms to anticipate future energy demand with impressive accuracy. Predicting how much power will be needed at any particular time is the aim of load forecasting, which requires this capacity. Precise load forecasting enables grid managers to optimize supply and demand, minimizing energy loss and decreasing the requirement for expensive peaking power plants, which are usually activated during spikes in demand. Furthermore, energy distribution can be optimized by machine learning. Machine learning models are able to detect inefficiencies in the grid's data and make recommendations for changes that will guarantee that energy is distributed as efficiently as feasible. This helps lower operating expenses and increase the stability of the energy supply in addition to increasing the grid's overall efficiency. Through predictive maintenance, machine learning is also a key factor in enhancing the dependability of smart grids. Conventional maintenance methods can be expensive and ineffective since they frequently rely on set timetables or reactive responses to equipment breakdowns. On the other hand, predictive maintenance makes predictions about the likelihood of equipment failure by analyzing sensor data and previous maintenance records using machine learning algorithms. Predictive maintenance lowers the risk of unplanned outages by enabling grid operators to arrange repairs at the most convenient times by identifying possible problems before they result in equipment failures. Over time, these preemptive measures save a substantial amount of money by reducing downtime and extending the life of vital infrastructure.

Machine learning can be used to identify abnormalities in the grid that might point to problems or inefficiencies in addition to forecasting equipment failures. An ML algorithm may, for example, identify anomalous patterns in energy flows or voltage levels that might indicate the existence of a defect. Machine learning facilitates quicker response times and helps to stop minor difficulties from becoming large ones by warning operators about these concerns in real time.  interruptions. Smart grids are more susceptible to hackers as they grow more digitalized and networked. To guarantee the stability and security of the electricity supply, the system must be shielded against these dangers. Because machine learning offers more sophisticated threat detection and response capabilities, it is essential for improving the cybersecurity of smart grids.

Network traffic and other data can be analyzed by ML algorithms to find trends that might point to a cyberattack. An anomalously high traffic flow from a certain source, for instance, could be picked up by an ML model and indicate that the grid is being attacked. Machine learning allows grid operators to respond swiftly and efficiently, reducing the impact of cyberattacks and stopping them from propagating throughout the grid by spotting these threats in real time.
Apart from identifying possible threats, machine learning may also be utilized to enhance the overall resilience of the grid by recognizing potential weaknesses and suggesting appropriate security steps to mitigate them. Protecting smart grids against a constantly changing array of cyberthreats requires a proactive approach to cybersecurity. One of the most important aspects of the continuous development of smart grid technologies is machine learning. Machine learning is contributing to the development of a more effective, dependable, and sustainable energy system by facilitating predictive maintenance, cybersecurity, integration of renewable energy sources, and data analysis. It is anticipated that as technology develops, its significance in smart grid applications will increase, opening the door for even more advancements in energy distribution and control. The energy industry has a bright future thanks to the combination of machine learning and smart grid technologies, as enhanced analytics and data-driven insights will shape the next wave of energy solutions.

\section{Conclusion}
The incorporation of machine learning (ML) into smart grids is a noteworthy advancement in augmenting the efficacy, dependability, and intelligence of energy infrastructures. However, in order to properly utilize this potential, a few obstacles must be overcome. Significant obstacles include issues with data quality, scalability, and real-time processing capabilities. Critical considerations that need to be carefully considered are also those of model interpretability, cybersecurity, privacy, and the requirement for regulatory adaptability. These obstacles are significant, but they are not insurmountable. Innovative solutions that lessen these restrictions can be produced through ongoing research and development as well as cooperative initiatives between business, academia, and government. To guarantee that ML can be implemented safely and successfully in the intricate and dynamic context of smart grids, these problems must be resolved. In conclusion, even though ML presents smart grids with revolutionary possibilities, it is critical to acknowledge and resolve the underlying difficulties. By doing this, we can fully utilize machine learning to propel the development of energy systems that are more intelligent, resilient, and sustainable.

\textbf{Funding Details:} his research is funded in part by NSF Grants No.
2306109, and DOEd Grant P116Z220008 (1). Any opinions,
findings, and conclusions expressed here are those of the
author(s) and do not reflect the views of the sponsor(s).

\begin{IEEEbiography}[{\includegraphics[width=1in,height=1.25in,clip,keepaspectratio]{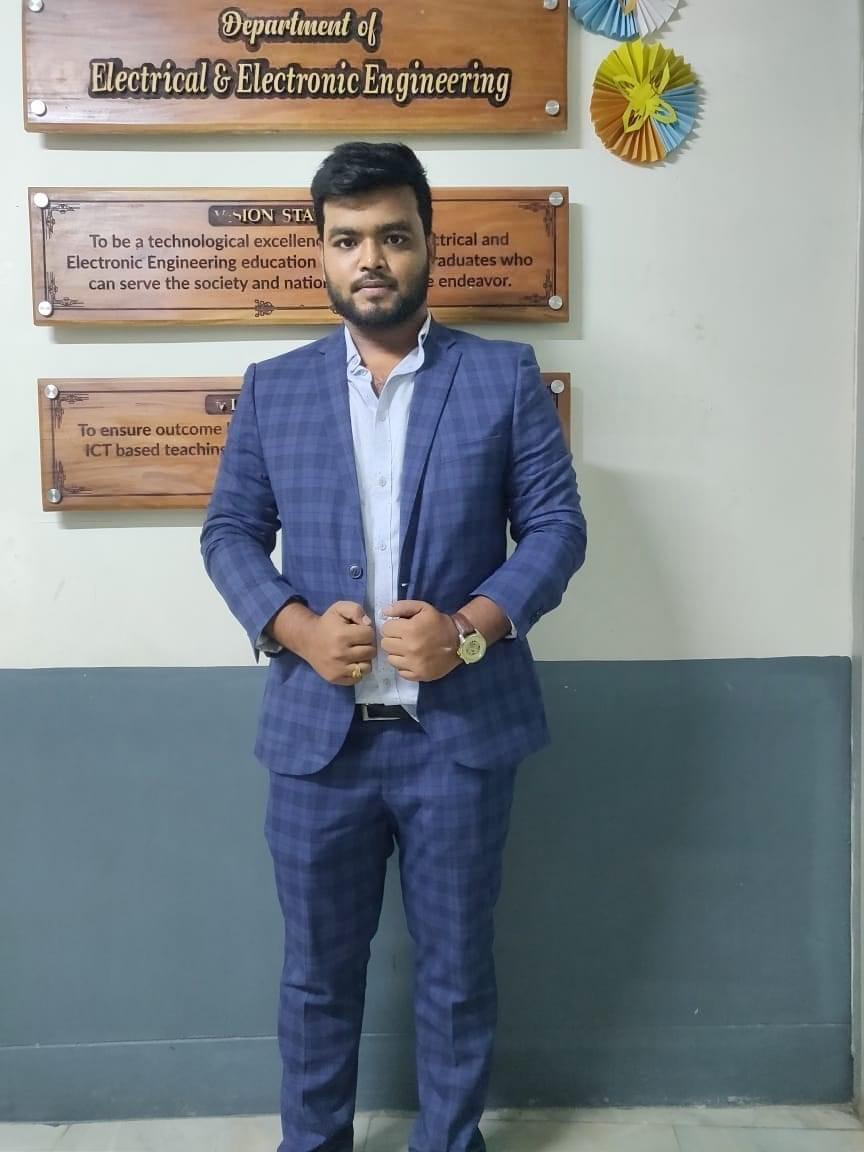}}]{Rashid Abdur} Abdur Rashid is currently an Electrical Engineer at Gazi Electric Co. since March 2021. He earned his BSc in Electrical and Electronic Engineering from Daffodil International University in January 2021, where he conducted research for his thesis titled "Biomass Resource Analyses and Future Bioenergy Scenarios." Additionally, he completed a Diploma in Engineering from MAWTS Institute of Technology in December 2016.
\end{IEEEbiography}

\begin{IEEEbiography}[{\includegraphics[width=1in,height=1.25in,clip,keepaspectratio]{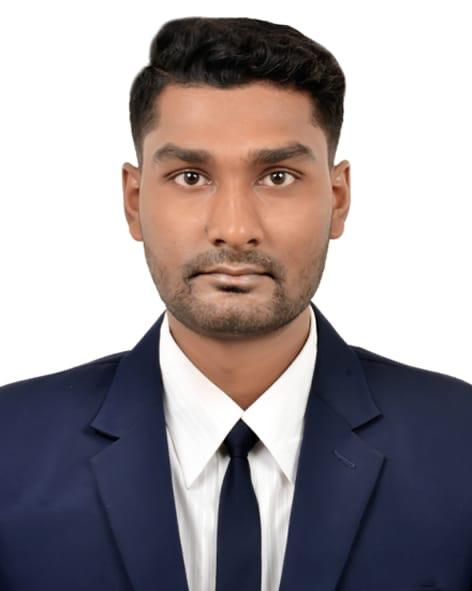}}]{BISWAS PARAG} Parag Biswas worked as an Officer (Electrical) in Supply Chain Management at Sajeeb Group from February 2021 to August 2022, where he developed and implemented supply chain strategies. He is currently pursuing a Master of Science in Engineering Management at Westcliff University (August 2022 - Present) and holds a Bachelor of Science in Electrical and Electronic Engineering from Daffodil International University, completed in January 2021, with a GPA of 3.15. Additionally, he conducted undergraduate research from July 2020 to January 2021, focusing on "Biomass Resource Analyses and Future Bioenergy Scenarios" under the supervision of Dr. Arnob Ghosh.
\end{IEEEbiography}

\begin{IEEEbiography}[{\includegraphics[width=1in,height=1.25in,clip,keepaspectratio]{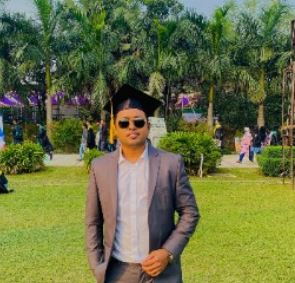}}]{MasumA bdullah Al} Abdullah Al Masum has a diverse background in software engineering and education. He completed his Master of Science in Information Technology from Westcliff University in May 2020. He is currently pursuing a Bachelor of Science in Computer Science at Daffodil International University, expected to graduate in November 2024. His professional experience includes working as a Junior Software Engineer at Intelle Hub Inc. from February 2021 to August 2022, an Instructor at Defence Care Academy from April 2020 to January 2021, and an intern in mobile application development at Grameen Solutions Limited from January to March 2020. Additionally, he has engaged in research and system development for the Defence Care Academy, focusing on analyzing and modifying existing software and constructing end-user applications. 
\end{IEEEbiography}

\begin{IEEEbiography}[{\includegraphics[width=1in,height=1.25in,clip,keepaspectratio]{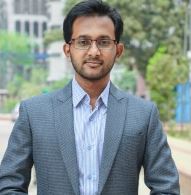}}]{MD Abdullah Al Nasim} received his Bachelor's degree in Computer Science and Engineering from Ahsanullah University of Science and Technology. His research interests include AI, Machine Learning, and Data Science, with a focus on practical applications and innovation. He has been recognized with awards such as the BASIS National ICT Awards, National Hackathon Winner, and an APICTA nomination. He is also the founder of several educational platforms that aim to enhance technical skills and promote STEM education.
\end{IEEEbiography}

\begin{IEEEbiography}[{\includegraphics[width=1in,height=1.25in,clip,keepaspectratio]{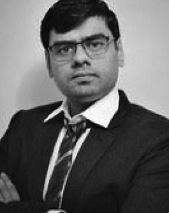}}]{ Gupta Kishor Datta}  The University of Memphis awarded Senior Member of IEEE a Ph.D. He is currently an Assistant Professor at Clark Atlanta University in Atlanta, Georgia's Department of Cyber-Physical Systems. Particularly in the area of adversarial machine learning, Dr. Gupta has made substantial contributions to the field, including one patent and multiple peer-reviewed papers. His areas of interest in study are computer vision, computer security, and bioinspired algorithms. As a member of the program committee for the Flagship Artificial Intelligence Conference AAAI-23, he is also actively involved in the academic community.

\end{IEEEbiography}

\EOD

\end{document}